\documentclass[11pt,a4paper]{article}
\usepackage[hyperref]{naaclhlt2019}

\usepackage{url}
\usepackage{array}
\usepackage{times}
\usepackage{amsmath}
\usepackage{amssymb}
\usepackage{graphicx}
\usepackage{multirow}

\aclfinalcopy
\def\aclpaperid{713}

\title{The Lower The Simpler: Simplifying Hierarchical Recurrent Models}

\author{
Chao Wang and Hui Jiang \\
Department of Electrical Engineering and Computer Science \\
Lassonde School of Engineering, York University \\
4700 Keele Street, Toronto, Ontario, Canada \\
{\tt $\{chwang, hj\}$@eecs.yorku.ca}
}

\date{}

\begin{document}
\maketitle
\begin{abstract}
To improve the training efficiency of hierarchical recurrent models without compromising their performance, we propose a strategy named as ``the lower the simpler'', which is to simplify the baseline models by making the lower layers simpler than the upper layers. We carry out this strategy to simplify two typical hierarchical recurrent models, namely Hierarchical Recurrent Encoder-Decoder (HRED) and R-NET, whose basic building block is GRU. Specifically, we propose Scalar Gated Unit (SGU), which is a simplified variant of GRU, and use it to replace the GRUs at the middle layers of HRED and R-NET. Besides, we also use Fixed-size Ordinally-Forgetting Encoding (FOFE), which is an efficient encoding method without any trainable parameter, to replace the GRUs at the bottom layers of HRED and R-NET. The experimental results show that the simplified HRED and the simplified R-NET contain significantly less trainable parameters, consume significantly less training time, and achieve slightly better performance than their baseline models.
\end{abstract}

\section{Introduction}
With the advance of various deep learning frameworks, neural network based models proposed for natural language understanding tasks are becoming increasingly complicated. To the best of our knowledge, a considerable part of these complicated models are both hierarchical and recurrent. For example, Hierarchical Recurrent Encoder-Decoder (HRED) \cite{sordonialessandro:2015, serbaniulianvlad:2016}, which is a conversational model, is constructed by stacking three layers of GRUs \cite{chokyunghyun:2014}. Besides, several well-known Machine Reading Comprehension (MRC) models, such as R-NET \cite{wangwenhui:2017} and FusionNet \cite{huanghsinyuan:2017}, are mainly composed of multiple layers of bidirectional GRUs (BiGRUs) or bidirectional LSTMs (BiLSTMs) \cite{hochreitersepp:1997}. The above hierarchical recurrent models have achieved excellent performance, but training them usually consumes a lot of time and memory, that is because their computational graphs contain a large amount of operators and trainable parameters, which makes their training computationally expensive. \\
According to \citet{williamsronaldj:1995}, in the training of recurrent neural networks, it is the backward propagation rather than the forward propagation that consumes the majority of the computational resources. Besides, considering the chain rule in the backward propagation, the complexity of computing gradients for a hierarchical recurrent model increases exponentially from the top layer of the model down to the bottom layer. Therefore, to improve the training efficiency of hierarchical recurrent models, our strategy is to \textbf{simplify the baseline models by making the lower layers simpler than the upper layers}, which we name as ``\textbf{the lower the simpler}''. Here ``simpler'' means containing less operators and trainable parameters. This strategy is guaranteed to work, since it can accelerate the computation of gradients, which is the substance of the backward propagation. However, there is still a big concern: once the baseline models are simplified, will their performance be compromised? \\
To address this concern, we carry out our proposed strategy to simplify two typical hierarchical recurrent models, namely HRED and R-NET, whose basic building block is GRU. Specifically, we propose Scalar Gated Unit (SGU), which is a simplified variant of GRU, and use it to replace the GRUs at the middle layers of HRED and R-NET. Besides, we also use Fixed-size Ordinally-Forgetting Encoding (FOFE) \cite{zhangshiliang:2015}, which is an efficient encoding method without any trainable parameter, to replace the GRUs at the bottom layers of HRED and R-NET. In the experiments, we separately compare the simplified HRED and the simplified R-NET with their baseline models in terms of both the training efficiency and the performance. The experimental results show that the simplified models contain significantly less trainable parameters, consume significantly less training time, and achieve slightly better performance than their baseline models.

\section{Baseline Models}
\subsection{Hierarchical Recurrent Encoder-Decoder}
Hierarchical Recurrent Encoder-Decoder (HRED) is a conversational model for building end-to-end dialogue systems. Since a dialogue is a sequence of sentences, where each sentence is a sequence of words, HRED models this hierarchy with a hierarchical recurrent structure. Specifically, HRED consists of three layers of GRUs, which from bottom to top separately serve as the sentence-level encoder, the dialogue-level encoder, and the decoder. The sentence-level encoder GRU iteratively takes the embeddings of the words in a sentence to update its hidden state, thus its final hidden state is a representation of the sentence. The dialogue-level encoder GRU iteratively takes the representations of the sentences in a dialogue to update its hidden state, thus its hidden state at each time-step is a representation of the current dialogue. The decoder GRU takes the current dialogue representation to initialize its hidden state so as to generate a response sentence word by word.

\subsection{R-NET}
R-NET is an end-to-end MRC model that predicts an answer span for each given passage-question pair. Specifically, R-NET consists of five layers, which from bottom to top are separately the embedding layer, the encoding layer, the matching layer, the self-matching layer, and the output layer. The embedding layer maps the words to the word-level embeddings and the character-level embeddings. The character-level embeddings are generated by processing the character embeddings of the words with a BiGRU and concatenating the forward GRU final hidden states and the backward GRU final hidden states. The encoding layer processes the concatenation of the word-level embeddings and the character-level embeddings with another BiGRU and concatenates the forward GRU outputs and the backward GRU outputs so as to generate the context representations. The matching layer uses a gated attention-based BiGRU to fuse the context representations of the question into those of the passage so as to generate the question-aware passage representations. The self-matching layer uses another gated attention-based BiGRU to fuse the question-aware passage representations into themselves so as to generate the final passage representations. On this basis, the output layer uses a pointer network \cite{vinyalsoriol:2015} to generate an answer span.

\section{Model Simplification}
\subsection{Scalar Gated Unit}
Just like LSTM, GRU is a recurrent structure that leverages gating mechanisms to capture long-term dependencies in sequential data:
\begin{align*}
& \text{Update Gate: } z_t = \sigma(W_z [h_{t-1}, x_t]) \\
& \text{Reset Gate: } r_t = \sigma(W_r [h_{t-1}, x_t]) \\
& \text{New Memory: } \hat{h}_t = tanh(W_h [r_t \odot h_{t-1}, x_t]) \\
& \text{Hidden State: } h_t = (1 - z_t) \odot h_{t-1} + z_t \odot \hat{h}_t
\end{align*}
Researchers have proposed several simplified variants of GRU. For example, \citet{zhouguobing:2016} proposed Minimal Gated Unit (MGU), which combines the update gate and the reset gate into a single forget gate. Compared with GRU, MGU contains less trainable parameters, consumes less training time, and achieves similar performance. However, in this paper, to better carry out our proposed ``the lower the simpler'' strategy, we propose Scalar Gated Unit (SGU), which is an even more simplified variant of GRU:
\begin{align*}
& \text{Scalar Update Gate: } z_t = \sigma(w_z [h_{t-1}, x_t]) \\
& \text{Scalar Reset Gate: } r_t = \sigma(w_r [h_{t-1}, x_t]) \\
& \text{New Memory: } \hat{h}_t = tanh(W_h [r_t * h_{t-1}, x_t]) \\
& \text{Hidden State: } h_t = (1 - z_t) * h_{t-1} + z_t * \hat{h}_t
\end{align*}
By comparing the formulation of SGU with that of GRU, it is easy to see that both the update gate $z_t$ and the reset gate $r_t$ change from the vectors in GRU to the scalars in SGU. Accordingly, the weights for generating the gates change from the matrices $W_z$ and $W_r$ in GRU to the vectors $w_z$ and $w_r$ in SGU. Besides, the gating operator also changes from the element-wise multiplication $\odot$ in GRU to the scalar multiplication $*$ in SGU. Therefore SGU is guaranteed to be the simplest among all the variants of GRU.

\subsection{Fixed-size Ordinally-Forgetting Encoding}
Fixed-size Ordinally-Forgetting Encoding (FOFE) is an encoding method that uses the following recurrent structure to map a varied-length word sequence to a fixed-size representation:
\begin{align*}
h_t =
\begin{cases}
0, & \text{if } t = 0 \\
\alpha * h_{t-1} + x_t, & \text{otherwise}
\end{cases}
\end{align*}
where $h_t$ is the hidden state at time step $t$, $x_t$ is the embedding of the $t$-th word, and $\alpha$ ($0 < \alpha < 1$) is the forgetting factor that decays the previous hidden state. Given a word sequence of length $N$, the final hidden state $h_N$ is a fixed-size representation of the word sequence. Although formulated as a recurrent structure, FOFE can actually be implemented with an efficient matrix multiplication. Besides, the forgetting factor $\alpha$ is designed as a hyper-parameter so that FOFE contains no trainable parameter. Therefore FOFE is guaranteed to be the simplest among all the recurrent structures. As for the performance, according to \citet{zhangshiliang:2015}, FOFE based language models outperform their LSTM based competitors.

\subsection{Simplified Models}
According to the above descriptions, SGU is simpler than GRU, and FOFE is simpler than SGU. Therefore, now we can carry out our proposed ``the lower the simpler'' strategy by using SGUs and FOFEs to replace certain GRUs in HRED and R-NET. For HRED, we keep the decoder GRU at the top layer unchanged, use a SGU to replace the dialogue-level encoder GRU at the middel layer, and use a FOFE to replace the sentence-level encoder GRU at the bottom layer. For R-NET, we keep the output layer, the self-matching layer, and the matching layer unchanged, use a bidirectional SGU (BiSGU) to replace the BiGRU that generates context representations at the encoding layer, and use a bidirectional FOFE (BiFOFE, i.e., running FOFE both forward and backward) to replace the BiGRU that generates character-level embeddings at the embedding layer. After conducting the above replacements, we finally obtain a simplified HRED and a simplified R-NET.

\section{Experiments}
\begin{table*}[t!]
\centering
\begin{tabular}
{m{0.11\linewidth}m{0.11\linewidth}m{0.15\linewidth}m{0.12\linewidth}m{0.15\linewidth}m{0.15\linewidth}}
\hline
\bf Model &
\bf Word \newline Embedding &
\bf Hidden States \newline (bottom-up) &
\bf Trainable \newline Parameters &
\bf Training Time \newline (secs * epochs) &
\bf Performance \newline (ppl, err rate) \\
\hline
\multirow{3}{0.1\linewidth}{Baseline HRED}
& 200 & 200-1200-200 & 10,777,003 & 4,100 * 33 & 35.72, 66.62\% \\
& 400 & 400-1200-400 & 18,740,403 & 4,660 * 29 & 34.35, 66.13\% \\
& 600 & 600-1200-600 & 28,223,803 & 5,700 * 29 & 34.11, 65.95\% \\
\hline
\multirow{3}{0.1\linewidth}{Simplified HRED}
& 200 & 200-1200-200 & 6,456,605 & 2,030 * 35 & 35.14, 66.46\% \\
& 400 & 400-1200-400 & 12,019,605 & 2,210 * 30 & 34.01, 66.05\% \\
& 600 & 600-1200-600 & 18,142,605 & 2,590 * 29 & 33.79, 65.89\% \\
\hline
\end{tabular}
\caption{\label{t1} Comparing the simplified HRED with the baseline HRED on MovieTriples.}
\end{table*}
\begin{table*}[t!]
\centering
\begin{tabular}
{m{0.11\linewidth}m{0.11\linewidth}m{0.15\linewidth}m{0.12\linewidth}m{0.15\linewidth}m{0.15\linewidth}}
\hline
\bf Model &
\bf Word \newline Embedding &
\bf Hidden States \newline (bottom-up) &
\bf Trainable \newline Parameters &
\bf Training Time \newline (secs * epochs) &
\bf Performance \newline (ppl, err rate) \\
\hline
Baseline HRED & 600 & 600-1200-600 & 40,231,401 & 51,770 * 33 & 46.29, 68.76\% \\
\hline
Simplified HRED & 600 & 600-1200-600 & 30,150,203 & 21,690 * 33 & 45.55, 68.55\% \\
\hline
\end{tabular}
\caption{\label{t2} Comparing the simplified HRED with the baseline HRED on Ubuntu.}
\end{table*}
\begin{table*}[t!]
\centering
\begin{tabular}
{m{0.25\linewidth}m{0.2\linewidth}m{0.2\linewidth}m{0.2\linewidth}}
\hline
\bf Model &
\bf Trainable \newline Parameters &
\bf Training Time \newline (secs * epochs) &
\bf Dev Performance \newline (EM / F1) \\
\hline
Baseline R-NET & 2,307,991 & 2454 * 41 & 71.1 / 79.5 \\
\hline
Simplified R-NET & 2,007,435 & 2085 * 38 & 71.2 / 79.7 \\
\hline
\end{tabular}
\caption{\label{t3} Comparing the simplified HRED with the baseline HRED on SQuAD.}
\end{table*}
\subsection{Experimental Datasets}
\textbf{Dialogue Datasets.} We compare the simplified HRED with the baseline HRED on two dialogue datasets, namely MovieTriples \cite{serbaniulianvlad:2016} and Ubuntu \cite{loweryanthomas:2017}. MovieTriples contains over $240,000$ dialogues collected from various movie scripts, with each dialogue consisting of three sentences. Ubuntu contains over $490,000$ dialogues collected from the Ubuntu chat-logs, with each dialogue consisting of seven sentences on average. Both MovieTriples and Ubuntu have been randomly partitioned into three parts: a training set ($80\%$), a development set ($10\%$), and a test set ($10\%$). \\
\textbf{MRC Dataset.} We compare the simplified R-NET with the baseline R-NET on an MRC dataset, namely SQuAD \cite{rajpurkarpranav:2016}. SQuAD contains over $100,000$ passage-question pairs with human-generated answer spans, where the passages are collected from Wikipedia, and the answer to each question is guaranteed to be a fragment in the corresponding passage. Besides, SQuAD has also been randomly partitioned into three parts: a training set ($80\%$), a development set ($10\%$), and a test set ($10\%$). Both the training set and the development set are publicly available, but the test set is confidential.

\subsection{Implementation Details}
\textbf{HRED.} We implement both the simplified HRED and the baseline HRED with TensorFlow \cite{abadimartin:2016}. For the word embeddings, we set their size to $200$, $400$, and $600$ on MovieTriples and $600$ on Ubuntu, initialize them randomly, and update them during the training. For the forgetting factor $\alpha$ of FOFE, we set it to $0.9$ on both MovieTriples and Ubuntu. For the hidden state size of the sentence-level encoder GRU, we set it to $200$, $400$, and $600$ on MovieTriples and $600$ on Ubuntu. For the hidden state size of the dialogue-level encoder GRU and SGU, we set it to $1200$ on both MovieTriples and Ubuntu. For the hidden state size of the decoder GRU, we set it to $200$, $400$, and $600$ on MovieTriples and $600$ on Ubuntu. For model optimization, we apply the Adam \cite{kingmadiederikp:2014} optimizer with a learning rate of $0.0001$ and a mini-batch size of $32$. For performance evaluation, we use both perplexity and error rate as evaluation metrics. \\
\textbf{R-NET.} We implement both the simplified R-NET and the baseline R-NET with TensorFlow. For the word-level embeddings, we initialize them with the $300$-dimensional pre-trained GloVe \cite{penningtonjeffrey:2014} vectors, and fix them during the training. For the character embeddings, we initialize them with the same pre-trained GloVe vectors, and update them during the training. For the forgetting factor $\alpha$ of FOFE, we set it to $0.7$. For the hidden state size of both the BiGRUs and the BiSGU, we set it to $128$. For model optimization, we apply the Adam optimizer with a learning rate of $0.0005$ and a mini-batch size of $32$. For performance evaluation, we use both Exact Match (EM) and F1 score as evaluation metrics, which are calculated on the development set.

\subsection{Experimental Results}
For model comparison in the training efficiency, we use the same hardware (i.e., Intel Core i7-6700 CPU and NVIDIA GeForce GTX 1070 GPU) to train both the baseline models and the simplified models. The experimental results show that our proposed ``the lower the simpler'' strategy improves the training efficiency of both HRED and R-NET without compromising their performance. On the one hand, as shown in Table~\ref{t1} and Table~\ref{t2}, the simplified HRED contains $25\%$--$35\%$ less trainable parameters, consumes over $50\%$ less training time, and achieves slightly better performance than the baseline HRED. Besides, Table~\ref{t1} also shows that appropriately scaling up the model brings better performance but consumes more resource, which implies that the simplified HRED will perform better than the baseline HRED when time or memory is limited. On the other hand, as shown in Table~\ref{t3}, the simplified R-NET contains $13\%$ less trainable parameters, consumes $21\%$ less training time, and achieves slightly better performance than the baseline R-NET.

\section{Conclusion}
In this paper, we propose a strategy named as ``the lower the simpler'', which is aimed at improving the training efficiency of hierarchical recurrent models without compromising their performance. This strategy has been verified on two typical hierarchical recurrent models, namely HRED and R-NET, where we replace their middle layers and bottom layers with two simpler recurrent structures. The significance of this paper lies in that it reveals a methodology for avoiding unnecessary complexity in training hierarchical recurrent models, which we believe is applicable to many other hierarchical recurrent models.

\section*{Acknowledgments}
This work is partially supported  by a research donation from iFLYTEK Co., Ltd., Hefei, China, and a discovery grant from Natural Sciences and Engineering Research Council (NSERC) of Canada.

\bibliography{naaclhlt2019}
\bibliographystyle{acl_natbib}

\end{document}